# Edge-Optimized Multimodal Learning for UAV Video Understanding via BLIP-2

Yizhan Feng[1,4], Hichem Snoussi[1,4], Jing Teng[2], Jian Liu[2], Yuyang Wang[2], Abel Cherouat[3,4] and Tian Wang[5]

[1] UR-LIST3N, University of Technology of Troyes, Troyes, France
E-mail: hichem.snoussi@utt.fr
[2] Institute of Artificial Intelligence, North China Electric Power University, Beijing, China
[3] UR-GAMMA3, University of Technology of Troyes, Troyes, France
[4] EUt+, Data Science Lab, European Union

[5] Institute of Artificial Intelligence, Beihang University, Beijing, China

**Abstract.** The demand for real-time visual understanding and interaction in complex scenarios is increasingly critical for unmanned aerial vehicles. However, a significant challenge arises from the contradiction between the high computational cost of large Vision language models and the limited computing resources available on UAV edge devices. To address this challenge, this paper proposes a lightweight multimodal task platform based on BLIP-2, integrated with YOLO-World and YOLOv8-Seg models. This integration extends the multi-task capabilities of BLIP-2 for UAV applications with minimal adaptation and without requiring task-specific fine-tuning on drone data.

Firstly, the deep integration of BLIP-2 with YOLO models enables it to leverage the precise perceptual results of YOLO for fundamental tasks like object detection and instance segmentation, thereby facilitating deeper visual-attention understanding and reasoning. Secondly, a content-aware key frame sampling mechanism based on K-Means clustering is designed, which incorporates intelligent frame selection and temporal feature concatenation. This equips the lightweight BLIP-2 architecture with the capability to handle video-level interactive tasks effectively. Thirdly, a unified prompt optimization scheme for multi-task adaptation is implemented. This scheme strategically injects structured event logs from the YOLO models as contextual information into BLIP-2's input. Combined with output constraints designed to filter out technical details, this approach effectively guides the model to generate accurate and contextually relevant outputs for various tasks.

Our platform achieves excellent zero-shot performance on UAV visual tasks without requiring additional specialized data. While maintaining core capabilities such as open-vocabulary object description and dynamic scene understanding, our solution significantly reduces computational and storage overhead. It provides a lightweight, efficient, and readily deployable multimodal interactive solution for resource-constrained UAV platforms.

# 1 Introduction

The evolution of visual language models (VLMs) has progressed from task-specific architectures to large-scale pretrained systems, and more recently, towards general-purpose multimodal agents. Early research focused on designing dedicated networks for tasks such as image captioning and visual question answering (VQA). A significant turning point emerged with the advent of transformer-based pretraining. Models like CLIP learned powerful image-text alignment representations through contrastive learning on large-scale web data, demonstrating remarkable zero-shot transfer capabilities [1]. Subsequent models, such as Flamingo, incorporated cross-attention mechanisms to enable efficient fusion of large language models (LLMs) with visual encoders, laying the groundwork for few-shot learning [2]. Closed-source models like GPT-4V and Gemini set industry benchmarks with their strong general understanding abilities. Meanwhile, the open-source community has produced notable models like InternVL [3], Qwen2-VL [4], and DeepSeek-VL2 [5], which continually push performance limits on specific benchmarks. Models such as VideoLLaMA 2 further extend VLM capabilities to video understanding by incorporating temporal modeling modules [6]. However, these advanced models typically rely on hundreds of billions of parameters to achieve their powerful capabilities, resulting in substantial computational and memory demands that hinder their deployment on resource-constrained edge devices [6].

The demand for real-time visual understanding and interaction in complex scenarios is increasingly critical for unmanned aerial vehicles (UAVs), with applications spanning infrastructure inspection, disaster response, precision agriculture, and low-altitude logistics [7]. These tasks require onboard intelligent systems to perform multi-modal tasks, such as object detection, dynamic scene understanding, VQA, and video report generation — under strict constraints of computational power, storage, and energy consumption. A significant contradiction exists between the hardware limitations of UAV edge devices and the high computational cost of large VLMs [8].

This challenge has spurred research into adapting VLMs for UAV applications. Existing approaches primarily follow two technical pathways: first, cloud-edge collaborative computing, which transmits visual data to the cloud for processing but is hampered by network latency and bandwidth, struggling to meet high real-time requirements; second, aggressive compression of existing large VLMs via pruning and quantization, which often severely compromises their core zero-shot generalization and open-vocabulary understanding capabilities[9,10]. There is a pressing need for a lightweight, general-purpose VLM platform that can natively and efficiently support multi-modal, multi-task operations on UAV edge devices [11].

Within this context, the CLIP and BLIP model families have garnered attention for their exceptional balance between efficiency and performance. CLIP's innovation lies in its dual-encoder architecture and contrastive learning objective, enabling it to learn from vast amounts of image-text pairs without fine-grained annotations, providing a powerful visual representation foundation for numerous tasks [12]. BLIP-2 represents an even more efficient pathway [13]. It introduces a lightweight Query Transformer

3(Q-Former) as a bridge to efficiently connect a frozen pretrained visual encoder with a frozen LLM. This design allows BLIP-2 to align visual features with the textual semantic space without costly end-to-end pretraining. By training only a minimal set of parameters, BLIP-2 demonstrates exceptional zero-shot performance across various Vision language tasks, making it an ideal foundational model for resource-constrained scenarios.

Despite its lightweight and efficient nature, applying BLIP-2 directly to UAV multi-modal tasks faces core limitations. Firstly, BLIP-2 is primarily optimized for static images and lacks inherent temporal modeling capabilities for video sequences, making it difficult to process continuous video streams from UAVs and perform cross-frame reasoning. Secondly, as a general-purpose model, its capability is weaker for tasks requiring high-precision spatial localization [14]. Fortunately, recent advancements like YOLO-World offer a powerful complementary solution [15]. As an open-vocabulary object detection model, YOLO-World maintains the high-speed, lightweight tradition of the YOLO series while enabling real-time detection and localization of arbitrary categories based on text prompts. Its precise structured output provides a solid perceptual foundation for VLMs [16].

Based on this analysis, targeted optimization and enhancement of BLIP-2 are urgently needed to achieve powerful and general-purpose multi-modal, multi-task capabilities on resource-constrained UAV platforms. The core objective of this work is to construct a lightweight multi-modal task platform for UAV edge devices by deeply integrating and optimizing BLIP-2 with YOLO-World, YOLOv8-Seg models with minimal engineering overhead. Our specific contributions are threefold:

- Efficient Model Integration and Capability Enhancement: Through deep integration of BLIP-2 with efficient YOLO-series models, we empower BLIP-2 to directly leverage the precise perceptual results of the latter, enhancing its visual understanding foundation without requiring time-consuming task-specific fine-tuning on UAV data.
- Innovation in Video-level Understanding: We design a content-aware keyframe sampling and feature fusion strategy, enabling the lightweight BLIP-2 model to evolve from static image understanding to handling video-level interactive tasks.
- Unified Prompt Engineering for Multi-Task Adaptation: We propose a prompt optimization scheme centered on event logs as context. By injecting the structured output from YOLO as contextual information, we guide the same model to generate accurate and relevant outputs for different tasks.

Our solution achieves excellent zero-shot performance on various UAV visual tasks without requiring additional specialized fine-tuning of BLIP-2. While maintaining core capabilities such as open-vocabulary description and dynamic scene understanding, it significantly reduces computational and storage overhead.

## 2  Methodology

The proposed lightweight multimodal task platform for UAVs is architected to deliver efficient, general-purpose video-level visual-language understanding in resource-constrained edge computing environments. This is achieved via the strategic integration of advanced pre-trained models with minimal architectural modifications. The overall platform architecture is illustrated in Fig. 1.

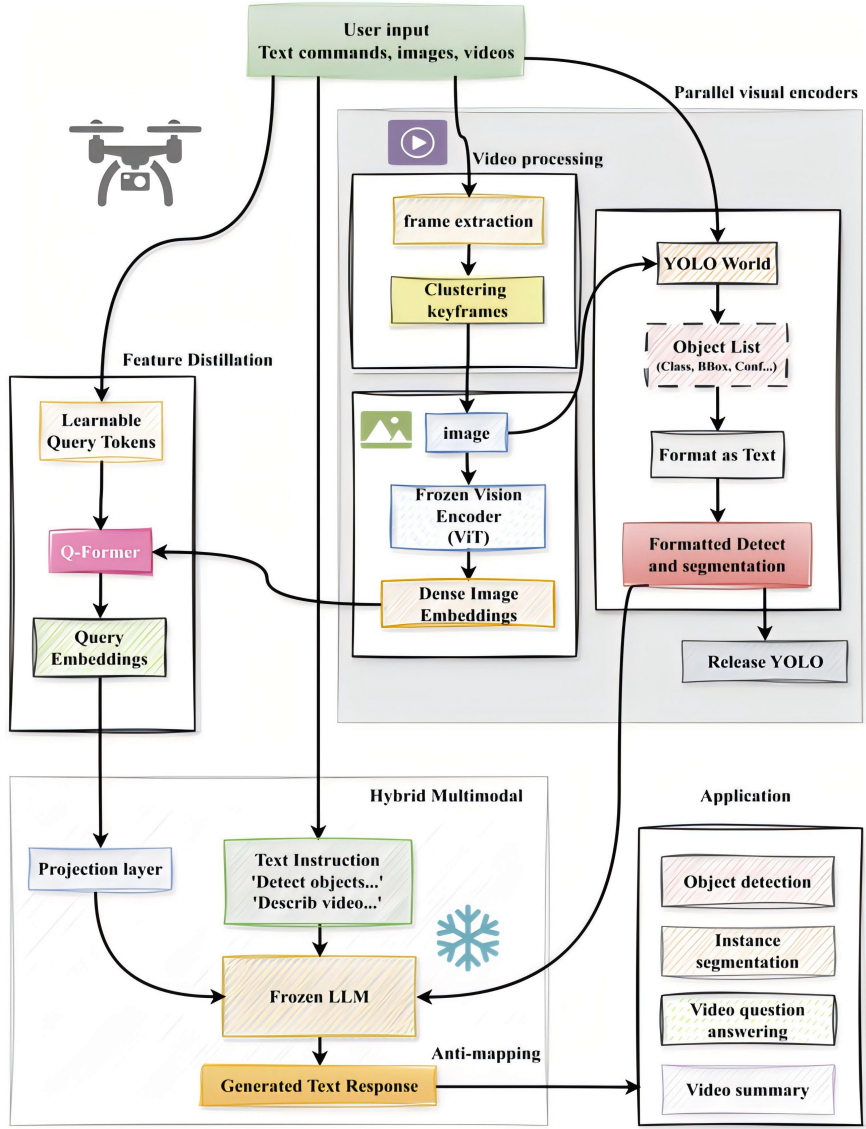

**Fig. 1.**  Architecture of the Proposed Multimodal Platform for UAV Video Understanding via BLIP-2 and YOLO Integration.

The overall workflow is as follows: First, the input video stream is fed into the visual perception module, which dynamically loads specialized models such as YOLO-World or YOLOv8-Seg to perform open-vocabulary object detection or instance segmentation on each frame, generating a structured event log containing object categories, locations, and frame numbers. Next, the video content representation module is activated. Its core is the visual-language model based on BLIP-2 [13]. This module does not process all video frames; instead, it employs a content-aware keyframe sampling strategy to intelligently select the most informative representative frames from the lengthy video sequence. The features of these keyframes are extracted via BLIP-2's visual encoder ViT and query transformer Q-Former and are further fused into a video-level representation imbued with spatiotemporal context. Finally, in the multimodal reasoning module, the fused video features, combined with carefully designed prompts, are input into the frozen large language model (LLM) for comprehension and reasoning. The final output dynamically generates multimodal responses tailored to the interactive task context.

To endow the platform with precise pixel-level or object-level perception capabilities, we deeply integrated advanced YOLO models. For open-vocabulary object detection tasks, we adopted the YOLO-World model, whose core innovation lies in proposing a Vision language path aggregation network RepVL-PAN [15]. RepVL-PAN injects textual semantic information into the image feature pyramid in the form of attention weights, guiding the model to focus on regions relevant to the text description. It utilizes multi-scale image features to enhance text embeddings, incorporating richer visual context . This enables YOLO-World to perform zero-shot object detection based on arbitrary category names or descriptive text provided by the user, without requiring retraining for these specific categories, significantly enhancing the platform's generalization capability.

For scenarios requiring finer pixel-level understanding, we integrated the YOLOv8-Seg model. Building upon object detection, this model adds an instance segmentation head capable of outputting precise pixel-level masks for targets. Both YOLO-World and YOLOv8-Seg maintain the high efficiency and low latency characteristic of the YOLO series, serving as the platform's "visual front-end" and providing structured perceptual results for subsequent advanced reasoning with BLIP-2.

The computational cost of processing all frames of a video directly is prohibitive. Therefore, we designed a content-aware keyframe sampling and feature fusion scheme, enabling BLIP-2 to evolve from static image understanding to possessing video-level contextual understanding capabilities. This scheme is a core innovation of this work. To overcome the limitations of simple uniform sampling, we employed a dynamic K-Means clustering algorithm to select the most informative representative frames from the video frames. The specific process is: First, perform moderately dense sampling on the input video subsequently, use BLIP-2's built-in visual encoder ViT and Q-Former to extract high-level semantic features for each sampled frame [8]; then, dynamically determine the number of clusters K based on the total video duration, ensuring each cluster represents a distinct video segment; finally, select the frame closest to the cluster center from each cluster as the representative frame for that cluster and sort them by timestamp. This method ensures that the selected small number of keyframes can maximally cover the main content changes and key scenes in the video, providing more representative input for subsequent analysis.

The core structure of BLIP-2 lies in its Q-Former. The Q-Former is a lightweight Transformer module containing a fixed set of learnable query vectors. These query vectors interact with the visual features output by the frozen image encoder through a cross-attention mechanism, thereby extracting the visual representations most informative for the text generation task from the original image [13]. Our scheme fully utilizes this characteristic: the selected K keyframes are sequentially input into the BLIP-2 model, and the Q-Former generates a fixed-length sequence of visual tokens for each frame. Subsequently, the visual tokens from all K frames are concatenated along the sequence length dimension, forming a long visual token sequence. This lengthy sequence encodes the spatiotemporal information of the entire video and serves as the fused video-level representation, which is input to the subsequent large language model. In this way, the language model can perform comprehensive reasoning based on cross-frame visual context, enabling a deep understanding of the dynamic video content.

To guide the integrated model to complete different tasks efficiently and accurately, and to suppress model "hallucinations," we designed a unified prompt engineering scheme. The core idea of this scheme is to construct an instruction template rich in contextual information for the model. In the prompts, we clearly define the model's task role and impose hard constraints on its output to ensure it conforms to the current task's requirements for text, images, or video. Simultaneously, we inject the structured event logs generated by the YOLO series models as key contextual information into the prompts. This design combines YOLO's precise perception capability with BLIP-2's powerful semantic understanding ability, enabling the model to not only "see" the image but also "know" the specific detection results when answering questions, thereby generating more grounded and accurate answers [8]. This scheme does not require modifying model parameters; merely by designing the input prompts, it can flexibly guide the same set of model parameters to adapt to various task output requirements.

## 3 Experiments

We present an empirical evaluation of our proposed lightweight multimodal task platform for UAVs through practical operational testing. In our experiment, we utilize authentic first-person view UAV video footage as test data to validate our platform's overall performance across target detection, instance segmentation , video question-answering, and video summarization tasks. Our system successfully completes an end-to-end processing pipeline: we begin by performing frame-by-frame analysis of the video stream to generate structured event logs, then extract representative frames through our content-aware keyframe sampling approach, and finally generate natural language descriptions by combining these with our carefully engineered prompts. We design our processing flow to ensure real-time performance while maintaining peak

GPU memory usage at approximately 17GB, demonstrating our platform's fundamental feasibility in resource-constrained environments.

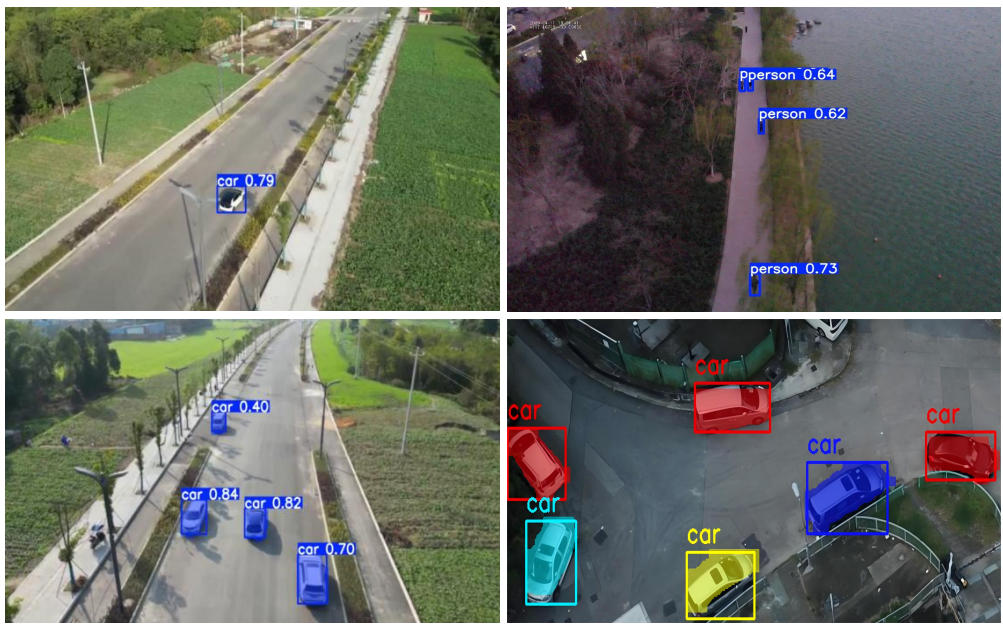

**Fig. 2.** The results of real-time object detection and instance segmentation on a UAV video sequence. The top two subfigures presents the outcome of real-time object detection, identifying vehicles and people on the road with bounding boxes. The bottom two subfigures demonstrates the result of instance segmentation, where each detected vehicle is precisely delineated at the pixel level, highlighted in color masking.

In assessing output quality, we find that our optimized prompts effectively guide the model to avoid technical details while yielding more natural visual commentary. For instance, our system describes vehicle detection results as "a car is detected on the road" rather than directly quoting coordinate data. The final generated video summary, "this is a video of a car driving on the road," accurately captures the core content. Our platform demonstrates significant advantages in temporal understanding as we enable the model to align event logs with video frame timestamps, generating descriptions tightly coupled to specific temporal points. As demonstrated in Fig. 2, we showcase the system's capability to perform real-time object detection and instance segmentation on a selected video frame through natural language interaction, highlighting its interactive multimodal functionality.

Table 1 reports the ablation comparison of different model variants in terms of functional capabilities (detection, segmentation, and video understanding) and system performance metrics (loading time, generation speed, memory footprint, and GPU memory usage. BLIP-2 serves as the performance and resource baseline for all variants. When only YOLO-World is introduced, the model acquires open-vocabulary object detection capability. The corresponding overhead is mainly reflected in a slight increase in memory and GPU consumption, while the loading time and overall resource usage do not exhibit any explosive growth. This suggests that the detection branch is structurally attached to the original vision–language backbone in a relatively independent and modular manner, exerting limited intrusion on the main inference pathway.

**Table 1.** Ablation Study of Model Capabilities and Performance Metrics.

| Model Variant | Det. | Seg. | Video | Load (s) | Speed (tok./s) | Mem. (GB) | GPU (GB) |
|---|---|---|---|---|---|---|---|
| BLIP-2 | ✗ | ✗ | ✗ | 14.97 | 29.12 | 14.66 | 14.68 |
| BLIP-2 + YOLO-World | ✓ | ✓ | ✗ | 17.25 | 28.84 | 15.62 | 15.65 |
| BLIP-2 + Frame Clustering | ✗ | ✗ | ✓ | 17.96 | 23.38 | 14.66 | 15.01 |
| Our Full Model | ✓ | ✓ | ✓ | 19.04 | 19.04 | 15.63 | 15.93 |

The variant with Frame Clustering exhibits the typical characteristics of video temporal modeling. Its generation speed decreases notably, indicating that multi-frame aggregation and temporal relationship modeling introduce additional computational paths and increase the cost of each forward pass. However, its memory and peak GPU usage remain very close to those of BLIP-2, implying that frame clustering enhances video understanding not through large-scale parameter inflation.

Our complete model, which integrates both detection and video-related functionalities, demonstrates only a marginal increase in resource consumption despite the expansion in supported capabilities. Under the current design, the aggregation of multi-task and multi-modal abilities does incur nontrivial time and memory overhead, but simultaneously achieves a deeper level of structural integration and engineering optimization.

**Table 2.** Zero-shot object detection evaluation on LVIS minival [15].

| Model | Backbone | Params(M) | AP | $AP_r$ | $AP_c$ | $AP_f$ |
|---|---|---|---|---|---|---|
| MDETR | R-101 | 169 | 24.2 | 20.9 | 24.3 | 24.2 |
| GLIP-T | Swin-T | 232 | 24.9 | 17.7 | 19.5 | 31 |
| GLIPv2-T | Swin-T | 232 | 26.9 | - | - | - |
| Grounding DINO-T | Swin-T | 172 | 25.6 | 14.4 | 19.6 | 32.2 |
| DetCLIP-T | Swin-T | 155 | 34.4 | 26.9 | **33.9** | 36.3 |
| YOLO-World-S | YOLOv8-S | 77 | 26.2 | 19.1 | 23.6 | 29.8 |
| YOLO-World-M | YOLOv8-M | 92 | 31.0 | 23.8 | 29.2 | 33.9 |
| **YOLO-World-L** | YOLOv8-L | 110 | **35.0** | **27.1** | 32.8 | **38.3** |

Table 2 presents the zero-shot object detection performance comparison on the LVIS minival [17]. AP (Average Precision) reflects the overall detection performance; $AP_r$, $AP_c$, and $AP_f$ denote detection accuracy for rare, common, and frequent categories, respectively, evaluating the model's generalization across objects of different frequencies.

The results demonstrate that our adopted YOLO-World-L achieves the highest AP with only 110M parameters while remaining highly competitive across other category-specific metrics. Compared to models using Swin-Transformer backbones [18], the YOLO-World series delivers superior performance with fewer parameters, highlighting its exceptional efficiency.

In implementing multi-turn interactive dialogues, we incorporate the complete YOLO event log as part of the conversational context. This design allows our model to maintain "long-term memory" of the entire video content when answering subsequent questions. Our approach effectively mitigates the inherent context length limitations and forgetting issues of large language models, enabling our platform to handle complex tasks requiring in-depth question-answering based on extended video content. We also monitor system resource consumption throughout our experiments. Through our dynamic model scheduling strategy, we successfully release the YOLO model after completing perception tasks, significantly reducing peak memory usage duration and reserving resource space for subsequent dialogue tasks.

## 4    Conclusion

This paper addresses the core challenge arising from the conflict between the high computational cost of large Vision language models and the limited computing resources available on UAV edge devices in complex scenarios demanding real-time visual understanding and interaction. The primary value of the proposed platform lies in its deep integration of the BLIP-2 model with YOLO-World and YOLOv8-Seg models, achieved with minimal engineering modifications. This integration successfully adapts powerful multimodal comprehension capabilities to resource-constrained UAV edge computing environments. Without requiring task-specific fine-tuning on dedicated drone data, our platform demonstrates excellent zero-shot performance on real-world UAV video data. It maintains core capabilities such as open-vocabulary object description and dynamic scene understanding, while significantly reducing computational and storage overhead through resource optimization strategies like dynamic model scheduling.

This research remains under ongoing development and refinement. Future work will explore integrating lightweight temporal modeling modules to enhance logical reasoning capabilities for long video sequences. Furthermore, constructing a high-quality and diverse evaluation dataset will be a key focus, aiming to facilitate objective and quantitative performance comparisons.

**Acknowledgements**

This work has been partially funded by BPI DreamScanner project.

# References

1. Radford A, Kim J W, Hallacy C, et al. Learning transferable visual models from natural language supervision[C]//International conference on machine learning. PmLR, 2021: 8748-8763.
2. Alayrac J B, Donahue J, Luc P, et al. Flamingo: a visual language model for few-shot learning[J]. Advances in neural information processing systems, 2022, 35: 23716-23736.
3. Chen Z, Wang W, Tian H, et al. How far are we to gpt-4v? closing the gap to commercial multimodal models with open-source suites[J]. Science China Information Sciences, 2024, 67(12): 220101.
4. Wang P, Bai S, Tan S, et al. Qwen2-vl: Enhancing Vision language model's perception of the world at any resolution[J]. arXiv preprint arXiv:2409.12191, 2024.
5. Wu Z, Chen X, Pan Z, et al. Deepseek-vl2: Mixture-of-experts Vision language models for advanced multimodal understanding[J]. arXiv preprint arXiv:2412.10302, 2024.
6. Cheng Z, Leng S, Zhang H, et al. Videollama 2: Advancing spatial-temporal modeling and audio understanding in video-llms[J]. arXiv preprint arXiv:2406.07476, 2024.
7. Limberg C, Gonçalves A, Rigault B, et al. Leveraging yolo-world and gpt-4v lmms for zero-shot person detection and action recognition in drone imagery[J]. arXiv preprint arXiv:2404.01571, 2024.
8. Li J, Li D, Xiong C, et al. Blip: Bootstrapping language-image pre-training for unified Vision language understanding and generation[C]//International conference on machine learning. PMLR, 2022: 12888-12900.
9. Feng Y, Snoussi H, Teng J, et al. Large Language Models to Enhance Multi-task Drone Operations in Simulated Environments[J]. Drones and Unmanned Systems, 2025: 212.
10. Feng Y, Snoussi H, Teng J, et al. Large language model-based multi-task UAVs-towards distilled real-time interactive control[C]//IET Conference Proceedings CP870. Stevenage, UK: The Institution of Engineering and Technology, 2023, 2023(39): 114-118.
11. Jin Y, Li J, Liu Y, et al. Efficient multimodal large language models: A survey[J]. arXiv preprint arXiv:2405.10739, 2024.
12. Radford A, Kim J W, Hallacy C, et al. Learning transferable visual models from natural language supervision[C]//International conference on machine learning. PmLR, 2021: 8748-8763.
13. Li J, Li D, Savarese S, et al. Blip-2: Bootstrapping language-image pre-training with frozen image encoders and large language models[C]//International conference on machine learning. PMLR, 2023: 19730-19742.
14. He K, Zhang X, Ren S, et al. Deep residual learning for image recognition[C]//Proceedings of the IEEE conference on computer vision and pattern recognition. 2016: 770-778.
15. Cheng T, Song L, Ge Y, et al. Yolo-world: Real-time open-vocabulary object detection[C]//Proceedings of the IEEE/CVF conference on computer vision and pattern recognition. 2024: 16901-16911.
16. Gu X, Lin T Y, Kuo W, et al. Open-vocabulary object detection via vision and language knowledge distillation[J]. arXiv preprint arXiv:2104.13921, 2021.
17. Kamath A, Singh M, LeCun Y, et al. Mdetr-modulated detection for end-to-end multi-modal understanding[C]//Proceedings of the IEEE/CVF international conference on computer vision. 2021: 1780-1790.
18. Liu Z, Lin Y, Cao Y, et al. Swin transformer: Hierarchical vision transformer using shifted windows[C]//Proceedings of the IEEE/CVF international conference on computer vision. 2021: 10012-10022.